\documentclass[5p,times,twocolumn]{elsarticle}

\usepackage{amsmath,amssymb,amsfonts,color}
\usepackage{graphicx}
\usepackage{textcomp}
\usepackage{xcolor}
\usepackage{enumerate}
\usepackage{setspace}
\usepackage{bm}
\usepackage{comment}
\usepackage{caption}
\usepackage{subcaption}
\usepackage{tipa}
\usepackage{footnote}
\usepackage{algorithm}
\usepackage{algorithmic}
\usepackage{url}
\usepackage{multirow}

\def\x{{\mathbf x}}

\usepackage{xcolor}

\usepackage{makecell}

\begin{document}


\begin{frontmatter}

\title{Privacy-Aware Continual Self-Supervised Learning on Multi-Window Chest Computed Tomography for Domain-Shift Robustness}

\author{Ren Tasai}
\ead{tasai@lmd.ist.hokudai.ac.jp}

\author{Guang Li}
\ead{guang@lmd.ist.hokudai.ac.jp}

\author{Ren Togo}
\ead{togo@lmd.ist.hokudai.ac.jp}

\author{Takahiro Ogawa}
\ead{ogawa@lmd.ist.hokudai.ac.jp}

\author{Kenji Hirata}
\ead{khirata@pop.med.hokudai.ac.jp}

\author{Minghui Tang}
\ead{toumeiki@hs.hokudai.ac.jp}

\author{Takaaki Yoshimura}
\ead{takaaki.ysm@med.hokudai.ac.jp}

\author{Hiroyuki Sugimori}
\ead{sugimori@hs.hokudai.ac.jp}

\author{Noriko Nishioka}
\ead{nnishioka@pop.med.hokudai.ac.jp}

\author{Yukie Shimizu}
\ead{yshimizu7298@med.hokudai.ac.jp}

\author{Kohsuke Kudo}
\ead{kkudo@med.hokudai.ac.jp}

\author{Miki Haseyama}
\ead{mhaseyama@lmd.ist.hokudai.ac.jp}

\address{Hokkaido University, \\ N-8, W-5, Kita-Ku, Sapporo, 060-0808, Japan}

\begin{abstract}
We propose a novel continual self-supervised learning (CSSL) framework for simultaneously learning diverse features from multi-window-obtained chest computed tomography (CT) images and ensuring data privacy.
Achieving a robust and highly generalizable model in medical image diagnosis is challenging, mainly because of issues, such as the scarcity of large-scale, accurately annotated datasets and domain shifts inherent to dynamic healthcare environments. Specifically, in chest CT, these domain shifts often arise from differences in window settings, which are optimized for distinct clinical purposes. Previous CSSL frameworks often mitigated domain shift by reusing past data, a typically impractical approach owing to privacy constraints.
Our approach addresses these challenges by effectively capturing the relationship between previously learned knowledge and new information across different training stages through continual pretraining on unlabeled images.
Specifically, by incorporating a latent replay-based mechanism into CSSL, our method mitigates catastrophic forgetting due to domain shifts during continual pretraining while ensuring data privacy. 
Additionally, we introduce a feature distillation technique that integrates Wasserstein distance--based knowledge distillation (WKD) and batch-knowledge ensemble (BKE), enhancing the ability of the model to learn meaningful, domain-shift-robust representations. 
Finally, we validate our approach using chest CT images obtained across two different window settings, demonstrating superior performance compared with other approaches.
\end{abstract}


\end{frontmatter}


\section{Introduction}
\label{sec.1}
Medical image analysis is crucial to clinical decision-making for diagnostic support. 
Its performance has been dramatically improved by the emergence of deep learning--based supervised learning (SL)~\cite{medical_DL_survey_2023, medical_DL_survey_2024, medical_DL_survey_2025}. 
While automating parts of the diagnostic process can enhance the quality and efficiency of clinical judgment, models deployed in critical medical settings must be designed to simultaneously achieve high accuracy and function robustly, with generalizability across diverse datasets and conditions.
However, SL is mainly limited by the significant shortage of large-scale, accurately annotated medical image datasets~\cite{SSL_anotation_1, SSL_anotation_2, SSL_anotation_3}, and this scarcity is exacerbated because annotating medical data, which must balance privacy protection with accuracy, requires extensive expertise and considerable effort.
Consequently, the model efficiency relies heavily on the availability of high-quality annotated data.
\par
As an approach for addressing this data-scarcity issue, self-supervised learning (SSL) has garnered attention~\cite{medical_SSL_survey_2024_1, medical_SSL_survey_2024_2, SSL_xray_li2022covid, li2022tri}. 
In SSL, a model is first pretrained using unlabeled data and then fine-tuned with a small amount of labeled data. 
This paradigm involves pretraining a model on unlabeled data and subsequently fine-tuning it with a small amount of labeled data. Additionally, reports reveal that SSL achieves outstanding performance while effectively reducing labeling costs~\cite{medical_SSL_survey_2025, li2023self, li2024rgmim}.
However, SSL still exhibits a key limitation: it lacks generalizability in real-world healthcare environments because the dynamic nature of clinical settings causes changes in the data distributions of medical images over time, resulting in domain shifts~\cite{domain_shift_1, domain_shift_2, domain_shift_3}.
This shift stems from differences across medical institutions, imaging equipment, and diagnostic objectives, resulting in high diversity across medical images. 
A prominent example is in chest computed tomography (CT), where images often comprise multiple domains, such as the mediastinal and lung window settings, each optimized for distinct clinical-observation purposes~\cite{multi_window_1, multi_window_2}. 
In these dynamic clinical settings, data with different characteristics arrive sequentially, necessitating continuous handling of domain shifts.
However, conventional SSL relies on a joint training scenario, where a model is trained only after collecting a large amount of unlabeled data. 
Consequently, the model cannot flexibly adapt to new data without expensive retraining~\cite{CSSL_cromo, CSSL_challenges}. 
Furthermore, satisfying this premise in real-world scenarios is often challenging owing to the high computational costs of retraining and strict privacy constraints~\cite{CSSL_cassle_Fini, CSSL_C2ASR}.
\par
In response to these challenges, continual SSL (CSSL) was recently applied to medical imaging~\cite{medical_CL_survey_1}. 
This approach involves allocating data with varying characteristics across multiple training stages for continual pretraining. Particularly, maintaining data-distribution diversity during this pretraining process enables the acquisition of rich feature representations that are beneficial for subsequent fine-tuning. 
Notably, CSSL mitigates data interference that typically occurs when integrating different modalities or diverse domains within joint SSL frameworks~\cite{medcoss, tasai_icassp_2025}.
Additionally, CSSL achieves good accuracy and generalizability compared to representative supervised continual learning (SCL) paradigms~\cite{CSSL_lump, CSSL_streaming_data_Hu}. 
Therefore, CSSL is projected to address labeling-cost reduction and dynamic-environment domain shifts.
\par
Notably, the primary challenge of CSSL is catastrophic forgetting~\cite{catastrophic_forgetting_1993, catastrophic_forgetting_2017}, which occurs when a model overwrites or forgets previously acquired knowledge while learning new concepts.
To mitigate this issue, CSSL conventionally adopts experience-replay-based approaches~\cite{experience_replay_1, experience_replay_2} from SCL. 
In this approach, a portion of the original images is stored in a memory buffer ($B$), enabling the model to retain and revisit past knowledge during later sequential training stages. 
This strategy is highly versatile, exhibiting applicability in a wide range of scenarios compared with other SCL approaches. However, in the medical-data context, the retention of past datasets is often complicated by privacy concerns~\cite{data_privacy_1, data_privacy_2}.
In the SCL field, latent replay (LR)-based approaches have been introduced to address catastrophic forgetting. These approaches achieve privacy preservation by storing the activations of intermediate layers in neural networks (NNs) and replaying them when learning new knowledge ~\cite{latent_replay_natural_1}. Specifically, they store feature representations instead of preserving the original images, leveraging them in subsequent learning stages to ensure data privacy.
Nevertheless, LR remains largely unexplored within the CSSL context, necessitating the exploration of a novel LR-based CSSL framework.
\par
To satisfy the aforementioned research gap, we propose a novel CSSL framework that simultaneously addresses domain shifts in dynamic environments while effectively mitigating catastrophic forgetting under privacy-constraint conditions.
The proposed framework maintains a $B$ that stores only the feature representations of past data, enabling continual pretraining of rich representations while preserving data privacy and distribution diversity.
To realize this, we develop a feature distillation method that integrates Wasserstein distance (WD)-based knowledge distillation (WKD) with a batch-knowledge ensemble (BKE).
While  WKD enforces distributional alignment between replayed and mini-batch features, BKE aggregates feature representations to enhance consistency and reduce domain interference.
This unified WKD--BKE design facilitates the learning of robust, generalizable features across multiple domains, exhibiting suitability for privacy-conscious continual learning (CL).
Further, to evaluate the effectiveness of the proposed framework, we pretrained a model using chest CT images acquired under two different window settings and evaluated its performance on two distinct public CT-image datasets. 
Extensive experiments demonstrated that the proposed framework consistently outperformed other approaches, achieving superior robustness and performance.
\par
The contributions of our study are summarized below. 
\begin{itemize}
    \item We propose a novel LR-based CSSL framework to ensure data privacy and effectively address catastrophic forgetting during pretraining with chest CT images across two domains.
    \item We introduce a novel WKD--BKE-integrated feature distillation method to simultaneously enable robust feature-representation learning and mitigate data interference.
    \item Our extensive experiments reveal that our method outperforms state-of-the-art approaches on two public chest-CT-image datasets.
\end{itemize}
\par
The remainder of this paper is organized as follows: Section 2 discusses the extant studies, Section 3 describes the details of the proposed CSSL framework, and Sections 4, 5, and 6 present the experiments, discussion, and conclusions, respectively.
\section{Related Studies}
\label{sec.2}
\subsection{Self-supervised learning for addressing domain shifts}
\label{2.1}
SSL has recently garnered significant attention in medical image analysis, which is characterized by limited annotated data. SSL has been applied across various modalities, including CT~\cite{SSL_CT_multi-domain_wolf2023sel, SSL_CT_multi-domain_jiang2025self,  tasai_gcce_2024}, magnetic resonance imaging (MRI)~\cite{SSL_MRI_multi-domain_chang2022self, SSL_MRI_multi-domain_fiorentino2024intensity}, and fundus imaging~\cite{SSL_fundus_multi-domain_Gu2023102904, SSL_fundus_multi-domain_mojab2020real}. 
However, domain shifts due to differences in imaging equipment, acquisition protocols, and diagnostic objectives typically compromise model reliability and robustness. To address this, multi-domain pretraining under the joint-training condition has been explored.
\par
In CT-based studies, Wolf et al.~\cite{SSL_CT_multi-domain_wolf2023sel} proposed a masked autoencoder (MAE)-based SSL method for convolutional NNs. The authors pretrained the model on a large-scale chest-CT dataset obtained from multiple medical institutions and demonstrated its effectiveness using classification tasks.
Similarly, Jiang et al.~\cite{SSL_CT_multi-domain_jiang2025self} investigated the robustness of SSL to domain shifts for tumor segmentation in non-small-cell lung cancer CT.
Employing MRI, Fiorentino et al.~\cite{SSL_MRI_multi-domain_fiorentino2024intensity} introduced an intensity-based self-supervised domain-adaptation approach for intervertebral disc segmentation. This approach effectively reduced annotation costs and improved generalizability across scanners with heterogeneous acquisition settings.
Mojab et al.~\cite{SSL_fundus_multi-domain_mojab2020real} employed fundus imaging to demonstrate the superior adaptability of SSL-pretrained models, which were trained on multi-device datasets to unseen domains in glaucoma detection.
\par
Overall, these studies demonstrated that SSL-based pretraining represents an effective strategy for mitigating domain shifts in medical image analysis.
However, studies revealed that representation learning across different modalities and domains can interfere with each other during pretraining, primarily because of their substantial differences, which ultimately results in data interference~\cite{medcoss, tasai_icassp_2025}. 
Furthermore, in the application of SSL to clinical settings that are characterized by dynamically changing data distributions, models often require retraining on the entire dataset. Such retraining requires considerable computational resources. Moreover, the often limited access to all data due to privacy constraints poses significant challenges for clinical applications.
\begin{figure*}[t]
    \centering
    \includegraphics[width=\linewidth]{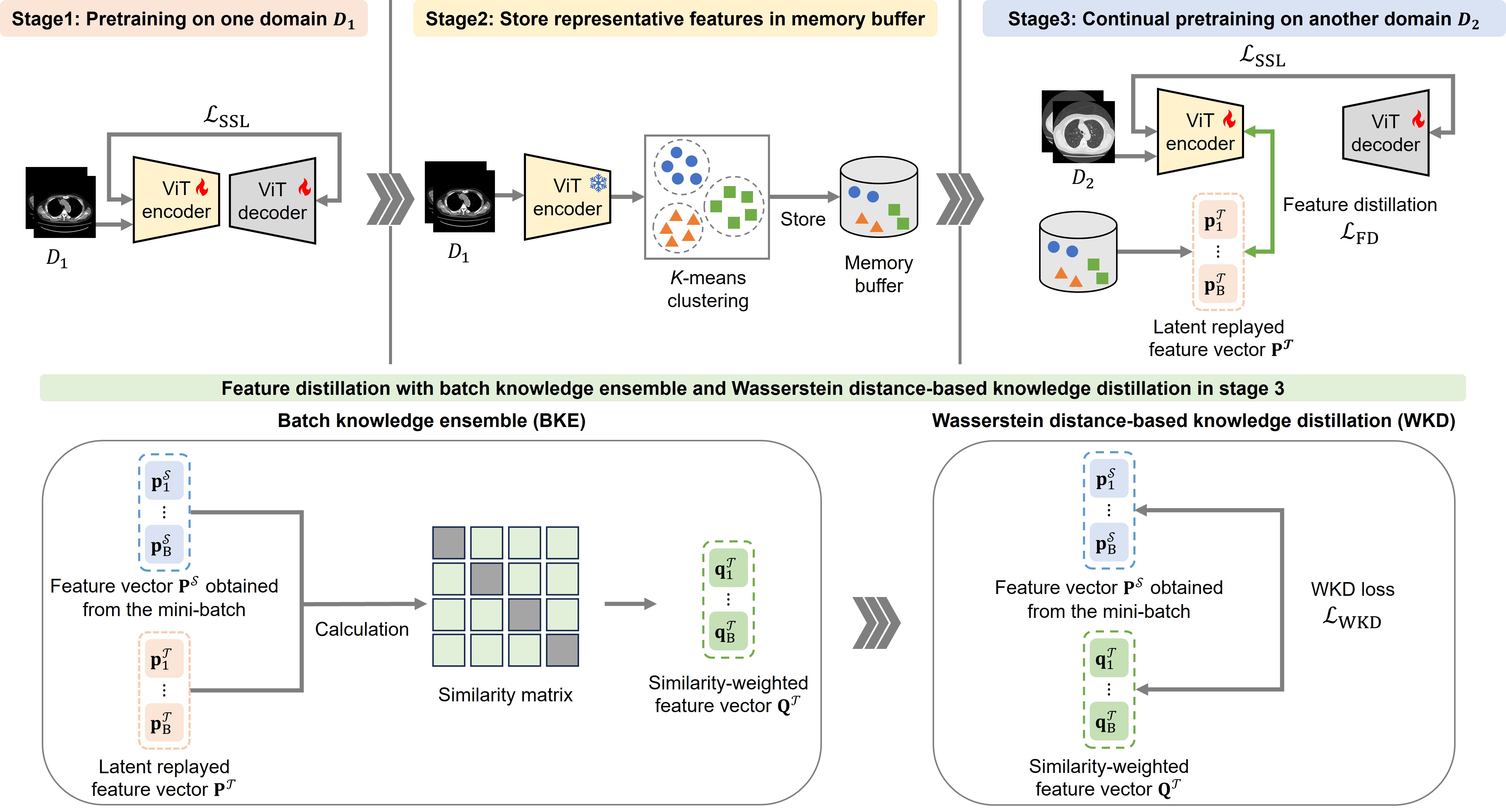} 
    \caption{Overview of the proposed continual self-supervised learning (CSSL) framework.} 
    \label{fig: method_overview}
\end{figure*}
\subsection{Continual self-supervised learning for addressing domain shifts}
\label{2.2}
In recent years, the application of CSSL has primarily focused on natural images, exploring its ability to handle incrementally arriving data. This capability is especially relevant in real-world settings where it is often impractical to assemble all data in advance.
Fini et al.~\cite{CSSL_cassle_Fini} experimented on DomainNet~\cite{DomainNet} involving sequential learning across multiple domains, demonstrating that CSSL outperformed major SSL and effectively addressed domain shifts. 
Furthermore, Hu et al.~\cite{CSSL_streaming_data_Hu} experimentally demonstrated on DomainNet that combining CSSL with simple SCL strategies, such as experience replay or parameter regularization~\cite{parameter_regularization}, can significantly mitigate performance degradation, even under substantial distributional shifts.
This and related studies provide valuable insights, collectively indicating that CSSL operates effectively in real-world scenarios, even in the presence of domain shifts.
\par
These advances motivated a growing interest in the application of CSSL to medical imaging analysis.
For instance, Ye et al.~\cite{medcoss} and Yao et al.~\cite{retcop} proposed methods for enhancing robustness and scalability in cross-modality learning. These approaches leverage experience replay and feature distillation~\cite{knowledge_distillation_survey_2025, knowledge_distillation_survey_2021} to efficiently integrate new modalities while suppressing the representational interference that often arises in conventional SSL.
However, these studies were limited to demonstrating effectiveness across multiple modalities and did not sufficiently examine applicability to multiple domains (i.e., domain shift within a single modality).
Conversely, a CSSL method was recently proposed~\cite{tasai_icassp_2025} using chest-CT images acquired under heterogeneous scanning conditions. The method involves the introduction of an experience-replay-based approach for balancing sample diversity and representativeness within $B$. This design enabled the acquisition of domain-invariant feature representations during continual pretraining. The authors deployed a COVID-19 classification task to demonstrate the effectiveness of CSSL in mitigating representational interference in conventional SSL while maintaining robustness across diverse domains.
\par
These medical-imaging studies generally adopted experience-replay-based approaches, which store past image samples in $B$ and reuse them in subsequent pretraining stages, thereby mitigating catastrophic forgetting.
Experience replay has demonstrated improved robustness in continual pretraining across multiple modalities and domains in medical imaging. 
Although this strategy is highly versatile and applicable to a wide range of scenarios compared with other CL methods, retaining past datasets is often impractical owing to concerns about preserving medical data privacy.
Moreover, only a few studies have explored continual pretraining within the CSSL framework under privacy-preserving constraints.
To address this gap, we propose a novel CSSL framework for chest-CT images spanning two domains, integrating an LR-based approach. 
Dissimilar to conventional experience replay, our method only retains feature representations in $B$ rather than the raw data, thereby achieving privacy-preserving continual pretraining while enabling the progressive acquisition of more expressive representations.
\section{Privacy-Aware Continual Self-Supervised Learning Integrating Latent Replay and Feature Distillation}
\label{sec.3}
Our CSSL framework comprises a three-stage sequential process for pretraining the encoder employed in subsequent downstream tasks.
In the first stage, SSL is performed using the initial dataset, $D_1$, from one chest-CT-image domain.
In the second stage, selected feature representations from $D_1$ are stored in $B$ to preserve data diversity and privacy.
In the third stage, SSL is performed again using the next (second-domain) dataset, $D_2$, from another domain. 
In this third stage, feature distillation involving WKD--BKE integration is performed using replayed features from $B$.
Afterward, fine-tuning is performed using labeled data.
Figure~\ref{fig: method_overview} shows an overview of the proposed CSSL framework.
\subsection{Stage 1: Self-supervised learning on the first-domain dataset}
\label{3.1}
The first pretraining stage proceeds with a model, $M_1$, using $D_1$. The MAE method~\cite{MAE} is employed to learn feature representations from the input data. During the learning, each image exhibiting $C$ channels is divided into $n$ patches of size $(V, V)$. Subsequently, a masking rate, $r$, is applied to randomly select $m = n \times r$ patches for masking. Next, these patches are converted into a sequence of tokens using a tokenizer, $\mathcal{T}_{M_1}$. 
Thereafter, the tokens corresponding to the $n-m$ unmasked patches are fed into the encoder $\phi_{M_1}$ to generate feature representations. The decoder, $\psi_{M_1}$, reconstructs the original masked patches, $X_m$, into reconstructed masked patches, $Y_m$, using the feature representations from the encoder along with the embeddings of $X_m$ from $\mathcal{T}_{M_1}$. Afterward, the model is optimized to minimize the mean squared error between $X_m$ and $Y_m$, as follows:
\begin{equation}
\label{eq:loss_ssl}
\mathcal{L}_{\text{SSL}} = \frac{1}{m \times V^{2}\times C} || Y_{m} - X_{m} ||^{2}_{2}.
\end{equation}
\par
This first stage terminates with the training of $M_1$ to capture comprehensive feature representations from $D_1$. 
This trained ($M_1$) model is subsequently employed in the third CSSL stage, which integrates $D_1$ and $D_2$.
\subsection{Stage 2: Sampling features in the memory buffer}
\label{3.2}
The second stage involves the selection of features stored in $B$. This process is crucial to capturing data-distribution changes across different stages and mitigating catastrophic forgetting. Additionally, the utilization of these selected features in the third stage ensures data privacy.  
\par
First, the feature representations of $D_1$ generated by the pretrained encoder, $\phi_{M_1}$, in the first stage are divided into $N \times \alpha$ clusters. Finally, $N \times \beta$ features closest to the cluster centers are selected and stored in $B$. In this algorithm, $N$ denotes the number of images in $D_1$, and $\alpha$ and $\beta$ are control parameters for the sampling ratio.
\subsection{Stage 3: Continual self-supervised learning with feature distillation using the second-domain dataset}
\label{3.3}
Following the SSL pretraining of model $M_1$ in the first stage, another model, $M_2$, is pretrained in the third stage using the MAE method. 
Notably, $M_2$ is trained using $D_2$ and the replayed features from $B$ in the second stage. 
Furthermore, WKD-BKE-integrated feature distillation enables $M_2$ to retain the knowledge acquired in the first stage while learning new representations for the second domain.
\subsubsection{Wasserstein distance--based knowledge distillation}
\label{3.3.1}
WKD facilitates knowledge retention by aligning the feature distributions of $M_2$ and $M_1$. 
Specifically, it compares the $M_1$-associated feature representations (replayed from $B$) with those generated by $M_2$ in the third stage.
\par
We consider a feature map derived from the feature representations, as follows: let the spatial height, width, and channel number of this feature map be $h$, $w$, and $l$, respectively.
Next, the feature map is transformed into a matrix, $\mathbf{F} \in \mathbb{R}^{l \times d}$, where $d = h \times w$, and the $i$-th column $\mathbf{f}_i \in \mathbb{R}^{l}$ represents the spatial features. 
Thereafter, we estimate the first- and second-order moments, $\boldsymbol{\mu} = \frac{1}{d} \sum_i \mathbf{f}_i$ and $\boldsymbol{\Sigma} = \frac{1}{d} \sum_i (\mathbf{f}_i - \boldsymbol{\mu})(\mathbf{f}_i - \boldsymbol{\mu})^T$, respectively, from these features. 
The feature distribution of the input images is modeled as a Gaussian distribution parameterized by the mean vector, $\boldsymbol{\mu}$, and covariance matrix, $\boldsymbol{\Sigma}$, as follows:
%
\begin{equation}
    \label{eq:wkd_1}
    \mathcal{N}(\boldsymbol{\mu}, \boldsymbol{\Sigma})=\frac{1}{|2 \pi \boldsymbol{\Sigma}|^{1/2}} \exp \{-\frac{1}{2}(\mathbf{f}-\boldsymbol{\mu})^T \boldsymbol{\Sigma}^{-1}(\mathbf{f}-\boldsymbol{\mu})\},
\end{equation}
%
where $|\cdot|$ is the matrix determinant.
Additionally, we define the teacher's and student’s feature distributions as $\mathcal{N}^{\mathcal{T}} \triangleq \mathcal{N}(\boldsymbol{\mu}^{\mathcal{T}}, \boldsymbol{\Sigma}^{\mathcal{T}})$ and $\mathcal{N}^{\mathcal{S}}$, respectively. 
The continuous WD between the two Gaussian distributions is expressed by the following:
\begin{equation}
    \label{eq:wkd_2}
    \mathrm{D}_{\mathrm{WD}}(\mathcal{N}^{\mathcal{T}}, \mathcal{N}^{\mathcal{S}})
    =\inf_q \int_{\mathbb{R}^l} \int_{\mathbb{R}^l}
    ||\mathbf{f}^{\mathcal{T}}-\mathbf{f}^{\mathcal{S}}||^2 
    q(\mathbf{f}^{\mathcal{T}}, \mathbf{f}^{\mathcal{S}}) 
    d\mathbf{f}^{\mathcal{T}} d\mathbf{f}^{\mathcal{S}},
\end{equation}
where inf represents the infimum, which is the greatest lower bound, 
$\mathbf{f}^{\mathcal{T}}$ and $\mathbf{f}^{\mathcal{S}}$ are Gaussian variables, and $||\cdot||^2$ denotes the Euclidean distance.
The joint distribution, $q$, is constrained such that its marginal distributions correspond to $\mathcal{N}^{\mathcal{T}}$ and $\mathcal{N}^{\mathcal{S}}$.
Thus, to minimize this equation and following~\cite{WKD_NeurIPS2024, wkd_eq}, we define the WKD loss function, $\mathcal{L}_{\text{WKD}}$, as follows:
\begin{equation}
    \label{eq:loss_wkd}
    \mathcal{L}_{\text{WKD}}
    =\gamma \mathrm{D}_{\text{mean}}(\boldsymbol{\mu}^{\mathcal{T}}, \boldsymbol{\mu}^{\mathcal{S}})+\mathrm{D}_{\operatorname{cov}}(\boldsymbol{\Sigma}^{\mathcal{T}}, \boldsymbol{\Sigma}^{\mathcal{S}}).
\end{equation}
Here, 
$D_{\text{mean}}(\boldsymbol{\mu}^{\mathcal{T}}, \boldsymbol{\mu}^{\mathcal{S}})= ||\boldsymbol{\mu}^{\mathcal{T}}-\boldsymbol{\mu}^{\mathcal{S}}||^2$ 
and 
$\mathrm{D}_{\operatorname{cov}}(\boldsymbol{\Sigma}^{\mathcal{T}}, \boldsymbol{\Sigma}^{\mathcal{S}}=
||\boldsymbol{\delta}^{\mathcal{T}}-\boldsymbol{\delta}^{\mathcal{S}}||^2$,
where $\boldsymbol{\delta}^{\mathcal{T}}$ and $\boldsymbol{\delta}^{\mathcal{S}}$ are the standard deviation vectors formed from the square root of the diagonal elements of $\boldsymbol{\Sigma}^{\mathcal{T}}$ and $\boldsymbol{\Sigma}^{\mathcal{S}}$, respectively.
Diagonal covariance matrices were employed for their robustness in estimating high-dimensional features as well as their computational efficiency~\cite{WKD_NeurIPS2024, wkd_diag}.
To balance the roles of the mean and covariance, we introduce a mean--covariance ratio hyperparameter, $\gamma$.
By computing $\mathcal{L}_{\text{WKD}}$, we enable the feature distribution of $M_2$ to align with that of $M_1$, thereby mitigating data interference due to inter-stage domain shifts (data-distribution differences across stages). 
\subsubsection{Batch knowledge ensemble}
\label{3.3.2}
We apply the BKE approach to enable $M_2$ to concurrently achieve robust learning while knowledge retention from the first stage.  
This approach enables feature distillation based on the similarity between $B$ feature representations, $\mathbf{P}^{\mathcal{T}}$, randomly replayed from the memory buffer and the feature representations, $\mathbf{P}^{\mathcal{S}}$, within the mini-batch generated by the encoder, $\phi_{M_2}$ of $M_2$, in the third stage.  
Therefore, knowledge is propagated and ensembled via the affinity between the feature representations replayed from $B$ and those generated by $M_2$ within mini-batches in the third stage.
\par
Let the batch size, number of tokens, and embedding dimension be $B$, $T$, and $E$, respectively.
Thus, $\mathbf{P}^{\mathcal{T}}, \mathbf{P}^{\mathcal{S}} \in \mathbb{R}^{B \times T \times E}$.
First, we obtain the similarity matrix, $\mathbf{A} \in \mathbb{R}^{B\times T \times T}$, by calculating the similarities between the replayed feature representations, $\{\mathbf{p}^{\mathcal{T}}_{1}, \dots, \mathbf{p}^{\mathcal{T}}_{B}\}$, retrieved from $B$, and the encoded visual features, $\{\mathbf{p}^{\mathcal{S}}_{1}, \dots, \mathbf{p}^{\mathcal{S}}_{B}\}$, extracted from a mini-batch of $B$ images, as follows:
\begin{equation}
    \label{eq:bke_1}
    \mathbf{A}_{i,j} = 
    (\hat{\mathbf{p}}^{\mathcal{T}\top}_{i} \hat{\mathbf{p}}^{\mathcal{S}}_{j}).
\end{equation}
In this equation, each feature representation is denoted as 
$\mathbf{p}^{\mathcal{T}}_{i}, \mathbf{p}^{\mathcal{S}}_{j} \in \mathbb{R}^{T \times E}$, 
where $\hat{\mathbf{p}}^{\mathcal{T}}_{i} = \mathbf{p}^{\mathcal{T}}_{i} / \Vert \mathbf{p}^{\mathcal{T}}_{i} \Vert_2$ 
and $\hat{\mathbf{p}}^{\mathcal{S}}_{j} = \mathbf{p}^{\mathcal{S}}_{j} / \Vert \mathbf{p}^{\mathcal{S}}_{j} \Vert_2$ 
represent the normalized feature representations.
The indices, $i$ and $j$, refer to the tokens within each mini-batch sample and replayed memory, respectively.
Further, we discard the diagonal entries using an identity matrix $\mathbf{I}$ to avoid self-knowledge reinforcement by $ \mathbf{A} = \mathbf{A} \odot (\mathbf{1}-\mathbf{I})$.
Next, we normalize $\mathbf{A} \in \mathbb{R}^{B\times T \times T}$ as follows:
\begin{equation}
    \hat{\mathbf{A}}_{i, j}=\frac{\exp \left(\mathbf{A}_{i, j}\right)}{\sum_{j \neq i} \exp \left(\mathbf{A}_{i, j}\right)}, \forall i \in\{1, \ldots, B\}.
\end{equation}
To prevent the excessive propagation and aggregation of noisy predictions, the optimized feature representation, $\mathbf{Q}$, is generated as a weighted sum of feature representation $\mathbf{P}^{\mathcal{T}}$ and propagated probability matrix $\hat{\mathbf{A}} \mathbf{P}^{\mathcal{T}}$, as follows:
\begin{equation}
    \mathbf{Q}=\omega \hat{\mathbf{A}} \mathbf{P}^{\mathcal{T}} + (1-\omega) \mathbf{P}^{\mathcal{T}}.
\end{equation}
Notably, propagation can proceed multiple times to generate $\mathbf{Q}$ for feature distillation:
\begin{equation}
\begin{split}
\mathbf{Q}_{(t)}
& = \omega\hat{\mathbf{A}}\mathbf{Q}_{(t-1)} + (1-\omega)\mathbf{P}^{\mathcal{T}},
\\ & = (\omega\hat{\mathbf{A}})^{t}\mathbf{P}^{\mathcal{T}} + (1-\omega)\sum_{i = 0}^{t-1}(\omega\hat{\mathbf{A}})^{i}\mathbf{P}^{\mathcal{T}},
\end{split}
\end{equation}
where $\omega$ is a weight factor and $t$ the $t$-th propagation and ensembling iteration.  
As the number of iterations approaches infinity, we obtain $\lim_{t \to \infty} (\omega \hat{\mathbf{A}})^t = 0$ and $\lim_{t \to \infty} \sum_{i=0}^{t-1} (\omega \hat{\mathbf{A}})^i = (\mathbf{I} - \omega \hat{\mathbf{A}})^{-1}$; hence we an approximate inference formulation can be obtained as follows:
\begin{equation}
    \label{eq:bke_5}
    \mathbf{Q}^\mathcal{T} = (1 - \omega)(\mathbf{I} - \omega \mathbf{\hat{A}})^{-1} \mathbf{P}^{\mathcal{T}}. 
\end{equation}
Thereafter, the optimized feature representation, $\mathbf{Q}^\mathcal{T}$, and $\mathbf{P}^{\mathcal{S}}$ are transformed into Eq. (2)--(3) to calculate $\mathcal{L}_{\text{FD}}$ = $\mathcal{L}_{\text{WKD}}$.
By computing the feature-distillation loss $\mathcal{L}_{\text{FD}}$, we facilitate the acquisition of robust feature representations while minimizing the deviation from those learned in the first stage.
\par
Next, our introduction of the LR-based approaches and WKD--BKE-integrated feature distillation into the CSSL framework enables the encoder to effectively capture the relationships between newly acquired data and previously learned knowledge.
This integration mitigates the effects of catastrophic forgetting during pretraining as well as facilitates the learning of richer, more robust feature representations.
Following the three-stage CSSL procedure, $\phi_{M_1}$ is fine-tuned on a separate labeled dataset for downstream tasks, such as classification.
During fine-tuning, $\phi_{M_1}$ is integrated with a randomly initialized task-specific multi-layer perceptron head and applied to the downstream tasks.
Algorithm 1 summarizes the proposed CSSL framework.
\begin{algorithm}[t]
    \caption{Algorithm of the proposed CSSL framework}
    \label{alg1}
    \begin{algorithmic}[1]
    \REQUIRE 
    $\{D_1, D_2\}$: two subsets from different domains, 
    $B$: memory buffer, 
    $\mathcal{T}_{M_1}, \mathcal{T}_{M_2}$: tokenizers, 
    $\phi_{M_1}, \phi_{M_2}$: encoders, 
    $\psi_{M_1}, \psi_{M_2}$: model-specific decoders, 
    $K\text{-}means(\cdot)$: $k$-means clustering operation, 
    $ClusterSample(\cdot)$: operation for sampling cluster centers, 
    $LatentReplay(\cdot)$: LR operation
    \ENSURE
    $\phi_{M_2}$, $\mathcal{T}_{M_2}$ 
    \\
    \textbf{Stage 1: SSL on $D_1$}
        \STATE Set the training dataset: $D \gets D_1$
        \STATE Update $\phi_{M_1}$, $\mathcal{T}_{M_1}$, and $\psi_{M_1}$ by minimizing $\mathcal{L}_{\text{SSL}}$, following Eq.~\eqref{eq:loss_ssl}
    \\
    \textbf{Stage 2: Sampling Features into the Memory Buffer}
        \STATE Obtain clusters: $C \gets K\text{-}Means(\phi_{M_1}(D_1))$
        \STATE Populate the memory buffer: $B \gets ClusterSample(C)$
    \\
    \textbf{Stage 3: CSSL with Feature Distillation on $D_2$}
        \STATE Set the training dataset: $D \gets D_2$
        \STATE Extract the mini-batch feature representations: $\mathbf{P}^\mathcal{S} \gets \phi_{M_2}(D_2)$
        \STATE Retrieve replayed feature representations from $B$: $\mathbf{P}^\mathcal{T} \gets LatentReplay(B)$
        \STATE Obtain $\mathbf{Q^\mathcal{T}}$ by calculating the similarity between $\mathbf{P}^\mathcal{S}$ and $\mathbf{P}^\mathcal{T}$, following Eq.~\eqref{eq:bke_1}--\eqref{eq:bke_5}
        \STATE Update $\phi_{M_2}$, $\mathcal{T}_{M_2}$, and $\psi_{M_2}$ 
               by minimizing $\mathcal{L}_{\text{SSL}}$ and $\mathcal{L}_{\text{FD}}$ 
               with $\mathbf{Q^\mathcal{T}}$ and $\mathbf{P}^\mathcal{S}$, following Eq.~\eqref{eq:loss_ssl} and \eqref{eq:loss_wkd}, respectively.
    \end{algorithmic}
\end{algorithm}
\section{Experiments}
\label{sec.4}
We comprehensively experiment on classification tasks to validate the effectiveness of the proposed CSSL framework. These experiments include ablation studies, hyperparameter analyses, and an investigation of the impact of extending pretraining stages.
The dataset and experimental settings are introduced in Subsection~\ref{4.1}.
Additionally, the classification-task performances with different pretraining datasets are discussed in Subsection~\ref{4.2}.
Furthermore, the impacts of hyperparameters on the experimental results are presented in Subsection~\ref{4.3}. 
The ablation Study of the proposed CSSL framework is discussed in Subsection~\ref{4.4}.
Finally, the impact of extending the pretraining stages in CSSL is demonstrated in Subsection~\ref{4.5}.
\subsection{Datasets and settings}
\label{4.1}
%
\begin{figure}[t]
    \centering
    \begin{subfigure}{0.3\linewidth}
        \centering
        \includegraphics[width=\linewidth]{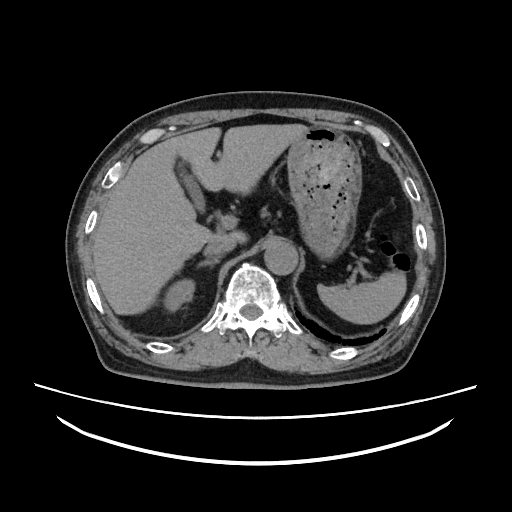}
        \caption{First-domain dataset ($D_1$)}
    \end{subfigure}
    \hspace{0.03\linewidth}
    \begin{subfigure}{0.3\linewidth}
        \centering
        \includegraphics[width=\linewidth]{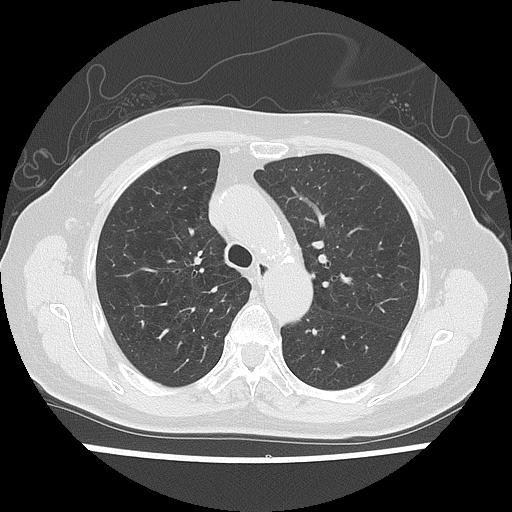}
        \caption{Second-domain dataset ($D_2$)}
    \end{subfigure}
    \caption{Examples of chest CT images on the subsets from the J-MID database: (a) First-domain dataset ($D_1$) and (b) Second-domain dataset ($D_2$).}
    \label{sample_images_J-MID}
\end{figure}
%
\begin{figure}[t]
    \centering
    \begin{subfigure}{0.3\linewidth}
        \centering
        \includegraphics[width=\linewidth]{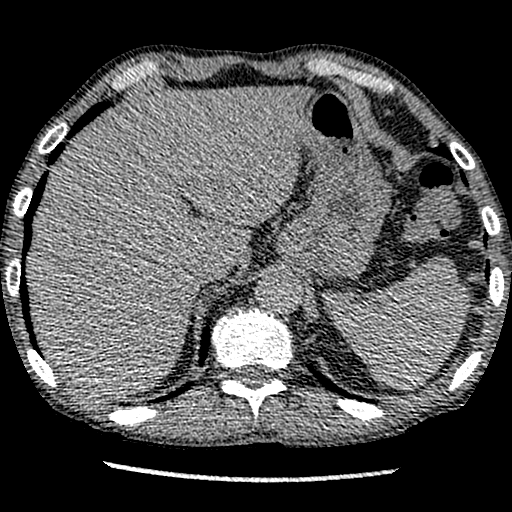}
        \caption{$D_1$}
    \end{subfigure}
    \hspace{0.03\linewidth}
    \begin{subfigure}{0.3\linewidth}
        \centering
        \includegraphics[width=\linewidth]{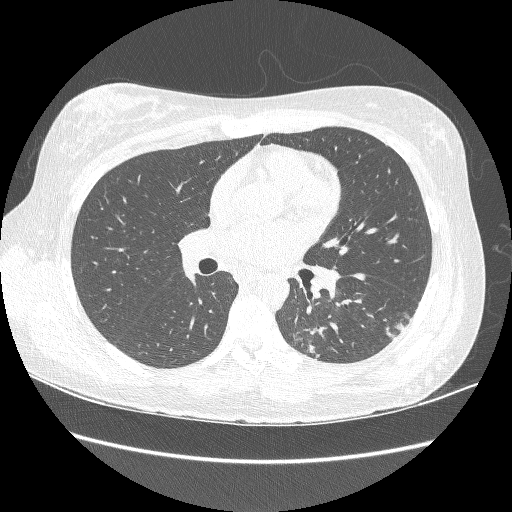}
        \caption{$D_2$}
    \end{subfigure}
    \caption{Examples of chest CT images on the subset from the RICORD dataset: (a) $D_1$ and (b) $D_2$.}
    \label{sample_images_RICORD}
\end{figure}
%
\begin{figure}[t]
    \centering
    \begin{subfigure}[t]{0.3\linewidth}
        \centering
        \includegraphics[width=\linewidth]{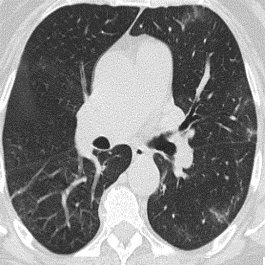}
        \caption{Corona virus disease 2019 (COVID-19)}
    \end{subfigure}
    \hspace{0.03\linewidth}
    \begin{subfigure}[t]{0.3\linewidth}
        \centering
        \includegraphics[width=\linewidth]{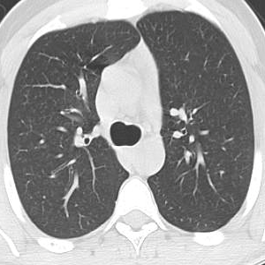}
        \caption{Normal}
    \end{subfigure}
    \caption{Examples of chest CT images on the SARS-CoV-2 CT-Scan Dataset: (a) COVID-19 and (b) Normal.}
    \label{SARS-CoV-2_CT_images}
\end{figure}
%
\begin{figure}[t]
    \centering
    \begin{subfigure}[t]{0.3\linewidth}
        \centering
        \includegraphics[width=\linewidth]{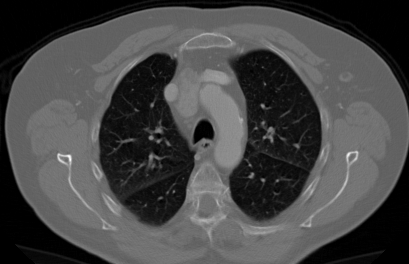}
        \caption{Adenocarcinoma}
    \end{subfigure}
    \hspace{0.03\linewidth}
    \begin{subfigure}[t]{0.3\linewidth}
        \centering
        \includegraphics[width=\linewidth]{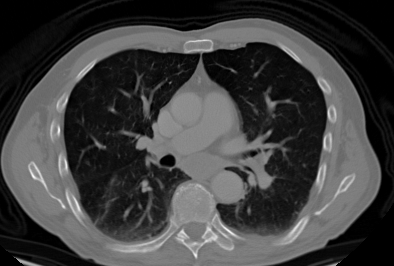}
        \caption{Large-cell carcinoma}
    \end{subfigure}
    \vspace{0.03\linewidth} \\
    %
    \begin{subfigure}[t]{0.3\linewidth}
        \centering
        \includegraphics[width=\linewidth]{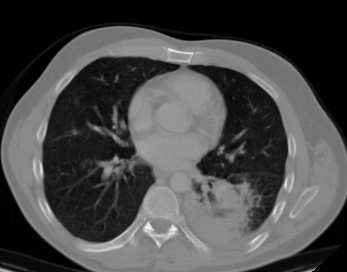}
        \caption{Squamous-cell carcinoma}
    \end{subfigure}
    \hspace{0.03\linewidth}
    \begin{subfigure}[t]{0.3\linewidth}
        \centering
        \includegraphics[width=\linewidth]{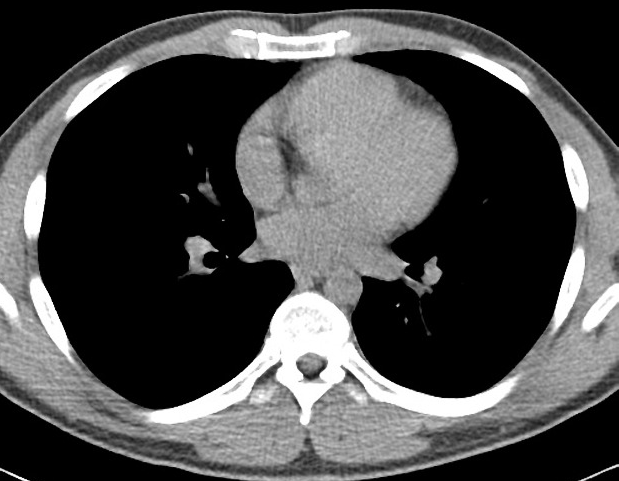}
        \caption{Normal}
    \end{subfigure}
    \caption{Examples of chest CT images from the Chest CT-Scan Images dataset: (a) adenocarcinoma, (b) large-cell carcinoma, (c) squamous-cell carcinoma, and (d) normal.}
    \label{Chest_CT_images}
\end{figure}
\begin{table*}[t]
    \centering
    \caption{Experimental results of the proposed method and conventional state-of-the-art methods pretrained on the J-MID subset.}
    \label{tab:main_result_J-MID}
    \begin{tabular}{lcc|ccc|ccc}
    \hline
    &   &   & \multicolumn{3}{c|}{SARS-CoV-2 CT-Scan Dataset~\cite{cov-2_dataset}} & \multicolumn{3}{c}{Chest CT-Scan images Dataset$^2$} \\
    \cline{4-9}
    Method & Domain &  & ACC & AUC & F1 & ACC & AUC & F1 \\
    \hline \hline
    Ours                        & \multirow{2}{*}{$D_1$ → $D_2$}    &
        & $\mathbf{0.873}_{\pm \mathbf{0.030}}$ & $\mathbf{0.953}_{\pm \mathbf{0.012}}$ & $\mathbf{0.873}_{\pm \mathbf{0.030}}$
        & $0.716_{\pm 0.042}$ & $0.943_{\pm 0.002}$ & $0.698_{\pm 0.054}$ \\
    MedCoSS~\cite{medcoss}      &                                   &
        & $0.858_{\pm 0.020}$ & $0.940_{\pm 0.016}$ & $0.858_{\pm 0.020}$
        & $0.662_{\pm 0.020}$ & $0.911_{\pm 0.008}$ & $0.642_{\pm 0.015}$ \\
    \hline
    Ours                        & \multirow{2}{*}{$D_2$ → $D_1$}    &
        & $0.777_{\pm 0.010}$ & $0.857_{\pm 0.016}$ & $0.777_{\pm 0.010}$
        & $\mathbf{0.833}_{\pm \mathbf{0.007}}$ & $\mathbf{0.965}_{\pm \mathbf{0.004}}$ & $\mathbf{0.845}_{\pm \mathbf{0.007}}$ \\
    MedCoSS~\cite{medcoss}      &                                   &
        & $0.763_{\pm 0.022}$ & $0.847_{\pm 0.024}$ & $0.761_{\pm 0.024}$
        & $0.822_{\pm 0.033}$ & $0.962_{\pm 0.009}$ & $0.833_{\pm 0.034}$ \\
    \hline
    \multirow{3}{*}{MAE~\cite{MAE}}
                                & $D_1$ + $D_2$ &
        & $0.784_{\pm 0.027}$ & $0.871_{\pm 0.025}$ & $0.783_{\pm 0.027}$
        & $0.807_{\pm 0.043}$ & $0.964_{\pm 0.008}$ & $0.804_{\pm 0.057}$ \\
                                & $D_1$         &
        & $0.729_{\pm 0.015}$ & $0.804_{\pm 0.014}$ & $0.724_{\pm 0.015}$
        & $0.665_{\pm 0.061}$ & $0.920_{\pm 0.003}$ & $0.668_{\pm 0.063}$ \\
                                & $D_2$         &
        & $0.760_{\pm 0.024}$ & $0.831_{\pm 0.041}$ & $0.756_{\pm 0.026}$
        & $0.519_{\pm 0.037}$ & $0.874_{\pm 0.011}$ & $0.509_{\pm 0.036}$ \\
    \hline
    Baseline                    & None          &
        & $0.620_{\pm 0.041}$ & $0.644_{\pm 0.050}$ & $0.599_{\pm 0.055}$
        & $0.495_{\pm 0.067}$ & $0.801_{\pm 0.023}$ & $0.500_{\pm 0.060}$ \\
    \hline
    \end{tabular}
\end{table*}
\begin{table*}[t]
    \centering
    \caption{Experimental results of the proposed method and conventional state-of-the-art methods pretrained on the RICORD dataset.}
    \label{tab:main_result_RICORD}
    \begin{tabular}{lcc|ccc|ccc}
        \hline
        &   &   & \multicolumn{3}{c|}{SARS-CoV-2 CT-Scan Dataset~\cite{cov-2_dataset}} & \multicolumn{3}{c}{Chest CT-Scan images Dataset$^2$} \\
        \cline{4-9}
        Method & Domain &  & ACC & AUC & F1 & ACC & AUC & F1 \\
        \hline\hline
        Ours                        & \multirow{2}{*}{$D_1$ → $D_2$}   &
            & $\mathbf{0.844}_{\pm \mathbf{0.034}}$ & $\mathbf{0.936}_{\pm \mathbf{0.023}}$ & $\mathbf{0.843}_{\pm \mathbf{0.034}}$
            & $0.689_{\pm 0.103}$ & $0.927_{\pm 0.031}$ & $0.672_{\pm 0.148}$ \\
        MedCoSS~\cite{medcoss}      &                                   &
            & $0.815_{\pm 0.045}$ & $0.911_{\pm 0.027}$ & $0.814_{\pm 0.046}$
            & $0.741_{\pm 0.061}$ & $0.952_{\pm 0.019}$ & $0.725_{\pm 0.079}$ \\
        \hline
        Ours                        & \multirow{2}{*}{$D_2$ → $D_1$}   &
            & $0.749_{\pm 0.020}$ & $0.821_{\pm 0.024}$ & $0.745_{\pm 0.021}$
            & $\mathbf{0.868}_{\pm 0.014}$ & $\mathbf{0.980}_{\pm 0.002}$ & $\mathbf{0.883}_{\pm 0.014}$ \\
        MedCoSS~\cite{medcoss}      &                                   &
            & $0.748_{\pm 0.027}$ & $0.814_{\pm 0.025}$ & $0.744_{\pm 0.029}$
            & $0.818_{\pm 0.059}$ & $0.973_{\pm 0.007}$ & $0.827_{\pm 0.062}$ \\
        \hline
        \multirow{3}{*}{MAE~\cite{MAE}}        
                                    & $D_1$ + $D_2$ &
            & $0.760_{\pm 0.026}$ & $0.844_{\pm 0.027}$ & $0.757_{\pm 0.030}$
            & $0.709_{\pm 0.047}$ & $0.940_{\pm 0.007}$ & $0.717_{\pm 0.054}$ \\
                                    & $D_1$         &
            & $0.727_{\pm 0.023}$ & $0.795_{\pm 0.007}$ & $0.722_{\pm 0.023}$
            & $0.614_{\pm 0.028}$ & $0.899_{\pm 0.016}$ & $0.632_{\pm 0.030}$ \\
                                    & $D_2$         &
            & $0.768_{\pm 0.008}$ & $0.843_{\pm 0.026}$ & $0.765_{\pm 0.010}$
            & $0.474_{\pm 0.056}$ & $0.854_{\pm 0.010}$ & $0.465_{\pm 0.063}$ \\
        \hline
        Baseline                    & None          &
            & $0.587_{\pm 0.019}$ & $0.611_{\pm 0.017}$ & $0.557_{\pm 0.030}$
            & $0.516_{\pm 0.037}$ & $0.799_{\pm 0.026}$ & $0.499_{\pm 0.061}$ \\
        \hline
    \end{tabular}
\end{table*}
For pretraining, we utilized a subset of the J-MID\footnote{\url{https://www.radiology.jp/j-mid/}} database, which contains large-scale CT scans from Japanese medical institutions, and the RICORD dataset~\cite{ricord_dataset}, an open dataset that was developed collaboratively by the Radiological Society of North America and international partners and contains chest CT scans collected from four countries. Each dataset was constructed with two domains based on mediastinal and lung window settings in chest CT images. Both domains are denoted as $D_1$ and $D_2$, and the labels are not used during pretraining. 
Specifically, for the J-MID subset, $D_1$ (the mediastinal window) contains 31,256 CT images, and  $D_2$ (the lung window) contains 26,403 CT images. The RICORD dataset comprises 12,897 $D_1$ (mediastinal window) images and 11,668 $D_2$ (lung window) images for pretraining. 
The corresponding images for each example are shown in Figs.~\ref{sample_images_J-MID} and ~\ref{sample_images_RICORD}.
For fine-tuning and evaluation, we utilized two public datasets: the SARS-CoV-2 CT-Scan Dataset~\cite{cov-2_dataset} and the Chest CT-Scan Images Dataset\footnote{\url{https://www.kaggle.com/datasets/mohamedhanyyy/chest-ctscan-images}}. Both datasets were used for the coronavirus disease 2019 (COVID-19) and chest-cancer classification tasks, respectively.
The data breakdown is as follows: the SARS-CoV-2 CT-Scan dataset comprises 787 training, 197 validation, and 250 test images, labeled into two (COVID-19 and Normal) classes.
The Chest CT-Scan Images Dataset comprises 490 training, 123 validation, and 315 test images labeled into four (adenocarcinoma, large-cell carcinoma, squamous-cell carcinoma, and normal) classes.
\par
In the pretraining of the MAE, the batch sizes were set to 64 and 32 in the first and second stages, respectively, and the masking ratio, $r$, was set to $0.75$, with ViT-B~\cite{vit} being deployed as the encoder. Next, augmentation techniques, such as random crop, resize, and flip,  were applied to the images. Additionally, a warm-up strategy was applied during the first 40 epochs, gradually increasing the learning rate from 0 to 0.00015. Subsequently, the learning rate was reduced to 0 via a cosine schedule. For \textit{k}-means sampling, the parameters, $\alpha$ and $\beta$, which determine the sampling ratio, were set to 0.01 and 0.05, respectively~\cite{medcoss}. Notably, $\gamma$, which adjusts the contributions of the mean and covariance in the WKD loss, $\mathcal{L}_{\text{WKD}}$, was set to 2.0 and 3.0 during the pretraining on the J-MID subset and RICORD dataset, respectively. In BKE, the hyperparameter, $\omega$, was set to 0.5~\cite{bke, SSL_xray_li2023boosting, li2022self}. The AdamW optimizer~\cite{adam} was utilized, with the learning rate set to 0.00005. Pretraining and fine-tuning were conducted for 300 and 80 epochs per stage, respectively. 
\par
For the evaluation metrics, we employed three metrics: two-class classification accuracy (ACC), the area under the receiver operating characteristic curve (AUC), and the F1-score (F1). To ensure robustness, we averaged the results across three of the four random seeds (0, 10, 100, and 1000). In all tables, the best performance is highlighted in bold for each experimental result. To evaluate the effectiveness of our method, we compared it with the following approaches: the state-of-the-art CSSL method for medical imaging, MedCoSS~\cite{medcoss}, MAE~\cite{MAE} simultaneously pretrained on $D_1$ and $D_2$, MAE pretrained only on $D_1$, and MAE pretrained only on $D_2$. As a baseline method, we employed a model that was fine-tuned without MAE-based self-supervised pretraining.
\subsection{Classification-task performance with different pretraining datasets}
\label{4.2}
Table~\ref{tab:main_result_J-MID} presents the classification results obtained after pretraining on the J-MID subset and evaluating on the SARS-CoV-2 CT-Scan dataset (for COVID-19 classification) and the Chest CT-Scan Images dataset (for lung cancer classification). Furthermore, to examine the effectiveness of the domain-pretraining order, we performed continual pretraining by interchanging domains $D_1$ and $D_2$.
Notably, the highest accuracy on the SARS-CoV-2 CT-Scan dataset was achieved when continual pretraining was performed from $D_1$ to $D_2$, whereas the best performance on the Chest CT-Scan Images dataset was obtained when the pretraining order was reversed from $D_2$ to $D_1$.
When pretrained on the same-order domains, the proposed method consistently outperformed MedCoSS across all evaluation metrics. 
This finding indicates that the proposed method effectively suppresses the catastrophic-forgetting effects.
Under the joint-learning scenario, the proposed method also surpassed the MAE approach, which conducts simultaneous pretraining on both domains, indicating that continual pretraining is more effective than simultaneous pretraining in mitigating domain interference.
\par
Table~\ref{tab:main_result_RICORD} also presents the results obtained after pretraining on the RICORD dataset.
Except for the continual pretraining order from $D_1$ to $D_2$, these results are consistent with those obtained using the J-MID database.
Collectively, the results demonstrate that the proposed method, which incorporates the LR-based approach, effectively mitigates data interference during continual pretraining.
\subsection{Impact of hyperparameters on the experimental results}
\label{4.3}
To investigate feature distillation for mitigating data interference and handling distributional differences across stages, we explored optimal parameter settings to minimize deviations from the knowledge acquired in the previous stage during subsequent learning.
Specifically, we examined the hyperparameter $\gamma$ in the WKD loss, $\mathcal{L}_{\text{WKD}}$, and the batch size of BKE to determine their optimal values.
The evaluation was on the COVID-19 classification task using the SARS-CoV-2 CT-Scan dataset.
\par
In the proposed method, the hyperparameter in the WKD loss represents $\gamma$, which controls the relative contributions of the mean and covariance terms.
Table~\ref{tab:ablation_result_gamma_J-MID} presents the classification results obtained with varying $\gamma$ values when pretraining was performed on the J-MID subset.
Table~\ref{tab:ablation_result_gamma_RICORD} presents the classification results with varying $\gamma$ values when pretraining was performed on the RICORD dataset.
For the proposed method, the optimal $\gamma$ setting was 2.0 for the $D_2$--$D_1$ continual pretraining order and 3.0 for the $D_1$ to $D_2$ order.
Accordingly, as $\gamma$ increases, the mean term in the WKD loss exerts more significant influence, indicating that the mean plays a more crucial role than the covariance.
\par
Next, in the BKE of the proposed method, knowledge is propagated and ensembled based on the affinity between the feature representations replayed from the $B$ and those generated by $M_2$ within mini-batches in the third learning stage.
Tables~\ref{tab:ablation_result_BS_J-MID} and Table~\ref{tab:ablation_result_BS_RICORD} present the classification results with varying batch sizes when pretraining was performed on the J-MID subset and RICORD dataset, respectively.
Notably, the optimal results were obtained with a batch size of 32 regardless of the utilized dataset.
\par
Overall, these findings indicate that the proposed method maintains robustness across a reasonable range of hyperparameter settings.
The observed performance-variation trends with respect to the mean--covariance balance and batch size are consistent and interpretable, indicating that the proposed framework behaves stably and predictably under different configurations.
This robustness demonstrates its practicality and reliability for continual pretraining across diverse medical-imaging domains.
%
\begin{table}[t]
    \centering
    \caption{Evaluation results on the SARS-CoV-2 CT-Scan dataset using the J-MID subset with varying hyperparameters, $\gamma$.
    Batch size was fixed at 32.}
    \label{tab:ablation_result_gamma_J-MID}
    \begin{tabular}{ccccc}
        \hline
        Domain & $\gamma$ & ACC & AUC & F1 \\
        \hline
        \multirow{5}{*}{$D_1$ → $D_2$}
            & 0.0 & $0.844_{\pm 0.015}$ & $0.932_{\pm 0.013}$ & $0.844_{\pm 0.015}$ \\
            & 1.0 & $0.852_{\pm 0.012}$ & $0.937_{\pm 0.017}$ & $0.852_{\pm 0.012}$ \\
            & 2.0 & $\mathbf{0.873}_{\pm \mathbf{0.030}}$ & $\mathbf{0.953}_{\pm \mathbf{012}}$ & $\mathbf{0.873}_{\pm \mathbf{0.030}}$ \\
            & 3.0 & $0.852_{\pm 0.009}$ & $0.939_{\pm 0.011}$ & $0.852_{\pm 0.009}$ \\
            & 4.0 & $0.840_{\pm 0.023}$ & $0.935_{\pm 0.019}$ & $0.840_{\pm 0.023}$ \\
        \cline{1-5}
        \multirow{5}{*}{$D_2$ → $D_1$}
            & 0.0 & $0.747_{\pm 0.016}$ & $0.843_{\pm 0.012}$ & $0.745_{\pm 0.016}$ \\
            & 1.0 & $0.752_{\pm 0.015}$ & $0.844_{\pm 0.027}$ & $0.751_{\pm 0.015}$ \\
            & 2.0 & $0.777_{\pm 0.010}$ & $0.857_{\pm 0.016}$ & $0.777_{\pm 0.010}$ \\
            & 3.0 & $0.774_{\pm 0.010}$ & $0.863_{\pm 0.008}$ & $0.774_{\pm 0.010}$ \\
            & 4.0 & $0.762_{\pm 0.021}$ & $0.849_{\pm 0.021}$ & $0.761_{\pm 0.015}$ \\
        \hline
    \end{tabular}
\end{table}
%
\begin{table}[t]
    \centering
    \caption{Evaluation results on the SARS-CoV-2 CT-Scan dataset for the model pretrained using the RICORD dataset with varying $\gamma$.
    Batch size was fixed at 32.}
    \label{tab:ablation_result_gamma_RICORD}
    \begin{tabular}{ccccc}
        \hline
        Domain & $\gamma$ & ACC & AUC & F1 \\
        \hline
        \multirow{5}{*}{$D_1$ → $D_2$}
            & 0.0 & $0.791_{\pm 0.041}$ & $0.913_{\pm 0.030}$ & $0.787_{\pm 0.042}$ \\
            & 1.0 & $0.804_{\pm 0.005}$ & $0.904_{\pm 0.014}$ & $0.803_{\pm 0.007}$ \\
            & 2.0 & $0.801_{\pm 0.050}$ & $0.901_{\pm 0.017}$ & $0.798_{\pm 0.055}$ \\
            & 3.0 & $\mathbf{0.844}_{\pm \mathbf{0.034}}$ & $\mathbf{0.936}_{\pm \mathbf{0.023}}$ & $\mathbf{0.843}_{\pm \mathbf{0.034}}$ \\
            & 4.0 & $0.832_{\pm 0.015}$ & $0.916_{\pm 0.011}$ & $0.831_{\pm 0.016}$ \\
        \cline{1-5}
        \multirow{5}{*}{$D_2$ → $D_1$}
            & 0.0 & $0.749_{\pm 0.008}$ & $0.822_{\pm 0.005}$ & $0.746_{\pm 0.009}$ \\
            & 1.0 & $0.743_{\pm 0.018}$ & $0.823_{\pm 0.018}$ & $0.742_{\pm 0.033}$ \\
            & 2.0 & $0.739_{\pm 0.012}$ & $0.820_{\pm 0.015}$ & $0.735_{\pm 0.015}$ \\
            & 3.0 & $0.749_{\pm 0.020}$ & $0.821_{\pm 0.024}$ & $0.745_{\pm 0.021}$ \\
            & 4.0 & $0.757_{\pm 0.019}$ & $0.833_{\pm 0.034}$ & $0.755_{\pm 0.020}$ \\
        \hline
    \end{tabular}
\end{table}
%
\begin{table}[t]
    \small
    \centering
    \caption{Evaluation results on the SARS-CoV-2 CT-Scan dataset for the model pretrained using the J-MID subset with varying batch sizes.
    $\gamma$ was fixed at 2.0.}
    \label{tab:ablation_result_BS_J-MID}
    \begin{tabular}{ccccc}
        \hline
        Domain & Batch Size & ACC & AUC & F1 \\
        \hline
        \multirow{4}{*}{$D_1$ → $D_2$} 
            & 16  & $0.839_{\pm 0.019}$ & $0.935_{\pm 0.008}$ & $0.839_{\pm 0.019}$ \\
            & 32  & $\mathbf{0.873}_{\pm \mathbf{0.030}}$ & $\mathbf{0.953}_{\pm \mathbf{0.012}}$ & $\mathbf{0.873}_{\pm \mathbf{0.030}}$ \\
            & 64  & $0.858_{\pm 0.027}$ & $0.933_{\pm 0.018}$ & $0.858_{\pm 0.027}$ \\
            & 128 & $0.865_{\pm 0.009}$ & $0.944_{\pm 0.008}$ & $0.865_{\pm 0.009}$ \\
        \cline{1-5}
        \multirow{4}{*}{$D_2$ → $D_1$} 
            & 16  & $0.748_{\pm 0.009}$ & $0.841_{\pm 0.006}$ & $0.745_{\pm 0.030}$ \\
            & 32  & $0.777_{\pm 0.016}$ & $0.857_{\pm 0.010}$ & $0.777_{\pm 0.010}$ \\
            & 64  & $0.743_{\pm 0.017}$ & $0.829_{\pm 0.006}$ & $0.741_{\pm 0.006}$ \\
            & 128 & $0.739_{\pm 0.010}$ & $0.825_{\pm 0.013}$ & $0.737_{\pm 0.013}$ \\
        \hline
    \end{tabular}
\end{table}
%
\begin{table}[t]
    \small
    \centering
    \caption{Evaluation results on the SARS-CoV-2 CT-Scan dataset for the model pretrained on the RICORD dataset with varying batch sizes.
    $\gamma$ was fixed at 3.0.}
    \label{tab:ablation_result_BS_RICORD}
    \begin{tabular}{ccccc}
        \hline
        Domain & Batch Size & ACC & AUC & F1 \\
        \hline
        \multirow{4}{*}{$D_1$ → $D_2$} 
            & 16  & $0.793_{\pm 0.037}$ & $0.868_{\pm 0.031}$ & $0.792_{\pm 0.038}$ \\
            & 32  & $\mathbf{0.844}_{\pm \mathbf{0.034}}$ & $\mathbf{0.936}_{\pm \mathbf{0.023}}$ & $\mathbf{0.843}_{\pm \mathbf{0.034}}$ \\
            & 64  & $0.832_{\pm 0.018}$ & $0.930_{\pm 0.010}$ & $0.832_{\pm 0.018}$ \\
            & 128 & $0.831_{\pm 0.031}$ & $0.918_{\pm 0.020}$ & $0.830_{\pm 0.031}$ \\
        \cline{1-5}
        \multirow{4}{*}{$D_2$ → $D_1$} 
            & 16  & $0.729_{\pm 0.005}$ & $0.802_{\pm 0.003}$ & $0.725_{\pm 0.003}$ \\
            & 32  & $0.749_{\pm 0.024}$ & $0.821_{\pm 0.020}$ & $0.745_{\pm 0.021}$ \\
            & 64  & $0.743_{\pm 0.013}$ & $0.813_{\pm 0.014}$ & $0.741_{\pm 0.013}$ \\
            & 128 & $0.750_{\pm 0.009}$ & $0.837_{\pm 0.015}$ & $0.748_{\pm 0.010}$ \\
        \hline
    \end{tabular}
\end{table}
\subsection{Ablation studies }
\label{4.4}
\begin{table}[t]
    \small
    \centering
    \caption{Results of the ablation studies on the latent replay (LR), Wasserstein distance (WD)-based knowledge distillation (WKD), and batch-knowledge ensemble (BKE) when the model was pretrained on the J-MID subset.}
    \label{tab:ablation_result_Novelty_J-MID}
    \begin{tabular}{ccc|ccc}
    \hline
    LR & WKD & BKE & ACC & AUC & F1 \\
    \hline
                   &        &        & $0.827_{\pm 0.020}$ & $0.935_{\pm 0.001}$ & $0.826_{\pm 0.021}$ \\
    \checkmark     &        &        & $0.833_{\pm 0.034}$ & $0.936_{\pm 0.016}$ & $0.832_{\pm 0.034}$ \\
    \checkmark     &        & \checkmark & $0.827_{\pm 0.024}$ & $0.929_{\pm 0.022}$ & $0.826_{\pm 0.024}$ \\
    \checkmark     & \checkmark &        & $0.845_{\pm 0.021}$ & $0.934_{\pm 0.018}$ & $0.845_{\pm 0.021}$ \\
    \checkmark     & \checkmark & \checkmark & $\mathbf{0.873}_{\pm \mathbf{0.030}}$ & $\mathbf{0.953}_{\pm \mathbf{0.012}}$ & $\mathbf{0.873}_{\pm \mathbf{0.030}}$ \\
    \hline
    \end{tabular}
\end{table}
\begin{table}[t]
    \small
    \centering
    \caption{Results of the ablation studies of LR, WKD, and BKE when the model was pretrained on the RICORD dataset.}
    \label{tab:ablation_result_Novelty_RICORD}
    \begin{tabular}{ccc|ccc}
        \hline
        LR & WKD & BKE & ACC & AUC & F1 \\
        \hline
                   &        &        & $0.804_{\pm 0.047}$ & $0.905_{\pm 0.019}$ & $0.803_{\pm 0.048}$ \\
    \checkmark     &        &        & $0.826_{\pm 0.052}$ & $0.906_{\pm 0.031}$ & $0.824_{\pm 0.054}$ \\
    \checkmark     &        & \checkmark & $0.840_{\pm 0.035}$ & $0.927_{\pm 0.023}$ & $0.839_{\pm 0.037}$ \\
    \checkmark     & \checkmark &        & $0.822_{\pm 0.047}$ & $0.906_{\pm 0.038}$ & $0.820_{\pm 0.050}$ \\
    \checkmark     & \checkmark & \checkmark & $\mathbf{0.844}_{\pm \mathbf{0.034}}$ & $\mathbf{0.936}_{\pm \mathbf{0.023}}$ & $\mathbf{0.843}_{\pm \mathbf{0.034}}$ \\
        \hline
    \end{tabular}
\end{table}
To evaluate the effectiveness of the proposed LR-based and WKD--BKE-integrated feature-distillation approaches, we performed an ablation study on the SARS-CoV-2 CT-Scan dataset, and
Tables~\ref{tab:ablation_result_Novelty_J-MID} and ~\ref{tab:ablation_result_Novelty_RICORD} present the results when pretraining was performed on the J-MID subset and RICORD dataset, respectively.
The first row reveals the performance of the baseline, which adopts an experience-replay-based approach with a \textit{k}-means sampling strategy as well as performs feature distillation using only the mean-squared-error loss.
The last row highlights the performance of the proposed method.
We confirmed that replacing the experience-replay-based approach with LR in the proposed CSSL framework improved classification accuracy, as LR eliminated the dependence on raw image storage by replaying latent representations, thereby reducing noise and redundancy that are inherent in pixel-level data. 
Consequently, the model retained informative and domain-invariant features more effectively, enhancing stability and knowledge retention during continual pretraining.
Furthermore, although integrating LR with only WKD or BKE did not yield significant improvement, its incorporation with both techniques yielded substantial performance improvements. 
Specifically, WKD aligns the feature distributions between past and newly acquired representations to suppress domain-specific biases.
Conversely, BKE exploits the similarity among feature representations within mini-batches and those replayed from $B$ to facilitate feature-level knowledge propagation as well as stabilize the optimization process.
Additionally, their integration offers complementary benefits, where WKD preserves consistency across domains, and BKE promotes coherence within batches.
This synergy enables the model to achieve more robust feature representations and enhanced classification accuracy.
\par
Thus, the newly introduced feature distillation enhances the knowledge-retention capability of the model while maintaining its new-information adaptability.
Overall, these results demonstrate the effectiveness of each component of the proposed method, underscoring their roles in mitigating data interference and improving continual-pretraining performance within the CSSL framework.
\subsection{Impact of stage extension on continual pretraining}
\label{4.5}
\begin{table}[t]
    \centering
    \caption{Experimental results for the extended pretraining stage. 
    Four-stage continual pretraining was conducted sequentially on RICORD and J-MID domains $D_1$ and $D_2$.}
    \label{tab: stage extension}
    \begin{tabular}{lccc}
        \hline 
        Stage & ACC & AUC & F1 \\
        \hline
        Stage 1 & $0.727_{\pm 0.023}$ & $0.795_{\pm 0.007}$ & $0.722_{\pm 0.023}$ \\
        Stage 2 & $0.844_{\pm 0.034}$ & $\textbf{0.936}_{\pm \textbf{0.023}}$ & $0.843_{\pm 0.034}$ \\
        Stage 3 & $0.754_{\pm 0.015}$ & $0.849_{\pm 0.021}$ & $0.754_{\pm 0.015}$ \\
        Stage 4 & $\textbf{0.857}_{\pm 0.020}$ & $\textbf{0.936}_{\pm \textbf{0.020}}$ & $\textbf{0.857}_{\pm \textbf{0.019}}$ \\
    \hline
    \end{tabular}
\end{table}
To examine the effect of progressive domain expansion during pretraining, we conducted four-stage continual pretraining on the RICORD dataset and J-MID subset, evaluating the resulting models on the SARS-CoV-2 CT-Scan dataset. The selected sequential training model was as follows: $D_1$--$D_2$ (RICORD), followed by $D_1$--$D_2$ (J-MID). This sequence was selected based on the results in Table~\ref{tab:main_result_J-MID} and Table~\ref{tab:main_result_RICORD}, which indicated that models pretrained on the RICORD dataset achieved lower accuracy on the SARS-CoV-2 CT-Scan task compared with those pretrained on the J-MID subset.
Therefore, we attempted to improve generalizability by first pretraining on the domains from the RICORD dataset before progressively expanding the pretraining to domains from the J-MID subset.
\par
Table~\ref{tab: stage extension} summarizes the results of this extended-pretraining experiment. 
In Stage 1, where the model was pretrained only on RICORD $D_1$, it exhibited limited performance on the SARS-CoV-2 CT-Scan classification task. 
In Stage 2, further pretraining on RICORD $D_2$ significantly improved all metrics, particularly the AUC, indicating enhanced representation robustness through pretraining on multiple domains within RICORD. 
In Stage 3, where the model was further trained on J-MID $D_1$, its performance decreased slightly owing to the domain characteristics, as this $D_1$ is less similar to the target SARS-CoV-2 CT-Scan dataset, whereas $D_2$ shares more common features with the target domain. 
In Stage 4, following the incorporation of J-MID $D_2$, the model recovered and exhibited improved performance.
\par
Overall, these findings indicate that extending the pretraining stages enables the model to learn more diverse and transferable representations, which consequently improve its performance on the downstream task.
These findings reveal that our approach can continually learn from images acquired from different domains across multiple medical institutions while preserving data privacy, thereby enabling more accurate, high-performance diagnostic capabilities.
\section{Discussion}
\label{sec.5}
Our experimental results demonstrated the effectiveness and robustness of our method across two distinct domains of unlabeled chest CT images, namely the mediastinal and lung windows, using two publicly available datasets (J-MID and RICORD). The proposed framework outperformed existing SSL and CSSL approaches, exhibiting notable advantages.
A major strength of the proposed method is its privacy-preserving design: the framework adopts LR, storing intermediate feature representations instead of original CT images. This strategy enables continual pretraining while eliminating the risk of sensitive data leakage, making the design suitable for dynamic clinical environments where imaging conditions and acquisition protocols change continually.
The integration of LR with WKD--BKE was crucial to achieving an adaptation--retention balance. Our ablation studies confirmed that LR mitigated catastrophic forgetting by simultaneously retaining essential latent features and maintaining privacy protection. 
Moreover, the synergy between WKD--BKE exerted complementary effects. WKD aligned feature distributions across different domains, reducing domain-specific bias, whereas BKE stabilized the training process by promoting feature-level knowledge propagation among mini-batches. 
Collectively, these mechanisms facilitated the learning of representations that were both domain invariant and discriminative, thereby improving performance in the downstream classification task.
\par
Despite these outstanding results, several challenges linger. 
A critical issue in continual pretraining is the significant dependence of model performance on the domain-exposure order: performance improves when the final domain is closely related to the evaluation data, decreasing otherwise. 
Therefore, for clinical applications, the relationship between the pretraining data used and the target images must be clarified before applying the model.
Another limitation of this study is its sole focus on chest CT images. Therefore, a key direction for future studies is to extend the framework to other medical-imaging modalities. 
Particularly, incorporating text information, such as physician-written diagnostic reports, may further improve generalizability through multi-modal representation learning. 
Additionally, while the proposed method stores only representative latent features in $B$, efficient memory management remains a challenge in long-term continual pretraining, as memory requirements increase inevitably with the increasing number of domains. Future studies may explore adaptive feature compression and dynamic memory-management techniques to maintain computational efficiency.
We plan to address these limitations via a more scalable CSSL framework capable of handling multi-modal, multi-institutional data. 
This essential extension will contribute to building a more generalizable and adaptive medical-imaging model that can be applied to diverse clinical settings.
\section{Conclusion}
\label{sec.6}
We proposed a privacy-aware CSSL framework to address the domain shifts in medical imaging.
The method incorporates an LR mechanism with WKD--BKE-integrated feature distillation, thus effectively mitigating catastrophic forgetting and preserving data privacy simultaneously.
Our experiments on multi-window chest-CT datasets demonstrated that our approach outperformed existing state-of-the-art SSL and CSSL methods, achieving superior robustness and generalizability across domains.
However, several limitations persisted, including the dependence of our approach on domain-exposure order, the efficiency of memory management, and the limited application of the framework to chest CT images.
We plan to address these limitations by extending the framework to multi-modal learning as well as integrating textual diagnostic information to establish a more scalable, generalizable, and privacy-preserving approach for medical artificial intelligence systems.
\section*{CRediT Authorship Contribution Statement}
Ren Tasai: Writing--review \& editing, Writing--original draft, Visualization, Validation, Software, Methodology, Data curation, Investigation, Formal analysis, Conceptualization. 
Guang Li: Writing--review \& editing, Supervision, Project administration, Methodology, Conceptualization, Formal analysis, Funding acquisition. 
Ren Togo: Writing--review \& editing, Supervision, Conceptualization, Funding acquisition. 
Takahiro Ogawa: Writing--review \& editing, Supervision, Conceptualization, Funding acquisition. 
Kenji Hirata: Writing--review \& editing, Data curation.
Minghui Tang: Writing--review \& editing, Data curation.
Takaaki Yoshimura: Writing--review \& editing, Data curation.
Hiroyuki Sugimori: Writing--review \& editing, Data curation.
Noriko Nishioka: Writing--review \& editing, Data curation.
Yukie Shimizu: Writing--review \& editing, Data curation.
Kohsuke Kudo: Writing--review \& editing, Data curation.
Miki Haseyama: Supervision, Funding acquisition.
\section*{Ethics in Publishing Statement}
This study was approved by the Research Ethics Committee, Faculty of Medicine, Juntendo University regarding the use of the J-MID database. The requirement for informed consent was waived by the committee because the study involved retrospective analysis of existing data and posed minimal risk to participants. All procedures were conducted in accordance with relevant ethical guidelines and regulations.
For the other datasets used (RICORD dataset, SARS-CoV-2 CT-Scan Dataset, and the Chest CT-Scan Images Dataset), no separate Ethics Statement was required, as this study involved no new experiments on human or animal subjects. Instead, it used publicly available or previously published datasets that were originally collected under appropriate ethical standards with necessary approvals, and contained no personally identifiable or sensitive information. Therefore, no additional ethical approval was needed for the present analysis.
\section*{Declaration of competing interest}
The authors declare no competing financial interests or personal relationships that could have influenced the work presented in this paper.
\section*{Acknowledgments}
This work was partly supported by JSPS KAKENHI Grant Numbers JP23K11141, JP23K21676, JP24K02942, JP24K23849, and JP25K21218.
We would like to thank the departments of radiology that provided the J-MID database, including Juntendo University, Kyushu University, Keio University, The University of Tokyo, Okayama University, Kyoto University, Osaka University, Hokkaido University, Ehime University, and Tokushima University.
\bibliographystyle{elsarticle-num}
\bibliography{ref}

\begin{thebibliography}{10}
\expandafter\ifx\csname url\endcsname\relax
  \def\url#1{\texttt{#1}}\fi
\expandafter\ifx\csname urlprefix\endcsname\relax\def\urlprefix{URL }\fi
\expandafter\ifx\csname href\endcsname\relax
  \def\href#1#2{#2} \def\path#1{#1}\fi

\bibitem{medical_DL_survey_2023}
Y.~Zhao, X.~Wang, T.~Che, G.~Bao, S.~Li, {Multi-task deep learning for medical image computing and analysis: A review}, Computers in Biology and Medicine 153 (2023) 106496.

\bibitem{medical_DL_survey_2024}
M.~E. Rayed, S.~S. Islam, S.~I. Niha, J.~R. Jim, M.~M. Kabir, M.~Mridha, Deep learning for medical image segmentation: {S}tate-of-the-art advancements and challenges, Informatics in Medicine Unlocked 47 (2024) 101504.

\bibitem{medical_DL_survey_2025}
O.~Shobayo, R.~Saatchi, Developments in deep learning artificial neural network techniques for medical image analysis and interpretation, Diagnostics 15~(9) (2025) 1072.

\bibitem{SSL_anotation_1}
F.~Garcea, A.~Serra, F.~Lamberti, L.~Morra, {Data augmentation for medical imaging: A systematic literature review}, Computers in Biology and Medicine 152 (2023) 106391.

\bibitem{SSL_anotation_2}
Y.~Li, J.~F. Wynne, Y.~Wu, R.~L. Qiu, S.~Tian, T.~Wang, P.~R. Patel, D.~S. Yu, X.~Yang, Automatic medical imaging segmentation via self-supervising large-scale convolutional neural networks, Radiotherapy and Oncology 204 (2025) 110711.

\bibitem{SSL_anotation_3}
P.~Singh, R.~Chukkapalli, S.~Chaudhari, L.~Chen, M.~Chen, J.~Pan, C.~Smuda, J.~Cirrone, {Shifting to machine supervision: Annotation-efficient semi and self-supervised learning for automatic medical image segmentation and classification}, Scientific Reports 14 (2024) 10820.

\bibitem{medical_SSL_survey_2024_1}
B.~VanBerlo, J.~Hoey, A.~Wong, A survey of the impact of self-supervised pretraining for diagnostic tasks in medical {X}-ray, {CT}, {MRI}, and ultrasound, BMC Medical Imaging 24~(1) (2024) 79.

\bibitem{medical_SSL_survey_2024_2}
X.~Zeng, N.~Abdullah, P.~Sumari, Self-supervised learning framework application for medical image analysis: {A} review and summary, BioMedical Engineering OnLine 23~(1) (2024) 107.

\bibitem{SSL_xray_li2022covid}
G.~Li, R.~Togo, T.~Ogawa, M.~Haseyama, {COVID-19} detection based on self-supervised transfer learning using chest {X}-ray images, International Journal of Computer Assisted Radiology and Surgery 18~(4) (2022) 715--722.

\bibitem{li2022tri}
G.~Li, R.~Togo, T.~Ogawa, M.~Haseyama, {TriBYOL}: {T}riplet {BYOL} for self-supervised representation learning, in: Proceedings of the IEEE International Conference on Acoustics, Speech and Signal Processing (ICASSP), 2022, pp. 3458--3462.

\bibitem{medical_SSL_survey_2025}
Z.~Zhao, L.~Alzubaidi, J.~Zhang, Y.~Duan, U.~Naseem, Y.~Gu, Robust and explainable framework to address data scarcity in diagnostic imaging, Computers in Biology and Medicine 197 (2025) 111052.

\bibitem{li2023self}
G.~Li, R.~Togo, T.~Ogawa, M.~Haseyama, Self-supervised learning for gastritis detection with gastric {X}-ray images, International Journal of Computer Assisted Radiology and Surgery 18~(10) (2023) 1841--1848.

\bibitem{li2024rgmim}
G.~Li, R.~Togo, T.~Ogawa, M.~Haseyama, {RGMIM}: {R}egion-guided masked image modeling for learning meaningful representations from {X}-ray images, in: Proceedings of the European Conference on Computer Vision (ECCV) Workshops, 2024, pp. 148--157.

\bibitem{domain_shift_1}
M.~Ceccon, D.~D. Pezze, A.~Fabris, G.~A. Susto, Multi-label continual learning for the medical domain: {A} novel benchmark, in: Proceedings of the IEEE/CVF Winter Conference on Applications of Computer Vision (WACV), 2025, pp. 7163--7172.

\bibitem{domain_shift_2}
W.~Li, Y.~Zhang, H.~Zhou, W.~Yang, Z.~Xie, Y.~He, {CLMS}: {B}ridging domain gaps in medical imaging segmentation with source-free continual learning for robust knowledge transfer and adaptation, Medical Image Analysis 100 (2025) 103404.

\bibitem{domain_shift_3}
X.~Liu, H.~A. Shih, F.~Xing, E.~Santarnecchi, G.~El~Fakhri, J.~Woo, Incremental learning for heterogeneous structure segmentation in brain tumor {MRI}, in: Proceedings of the International Conference on Medical Image Computing and Computer Assisted Intervention (MICCAI), 2023, pp. 46--56.

\bibitem{multi_window_1}
Q.~Wang, X.~Tan, L.~Ma, C.~Liu, Dual windows are significant: {L}earning from mediastinal window and focusing on lung window, in: Proceedings of the CAAI International Conference on Artificial Intelligence (CICAI), 2022, pp. 191--203.

\bibitem{multi_window_2}
H.~Lu, J.~Kim, J.~Qi, Q.~Li, Y.~Liu, M.~B. Schabath, Z.~Ye, R.~J. Gillies, Y.~Balagurunathan, Multi-window {CT} based radiological traits for improving early detection in lung cancer screening, Cancer Management and Research (2020) 12225--12238.

\bibitem{CSSL_cromo}
E.~Mushtaq, D.~N. Yaldiz, Y.~F. Bakman, J.~Ding, C.~Tao, D.~Dimitriadis, S.~Avestimehr, {CroMo-Mixup}: {A}ugmenting cross-model representations for continual self-supervised learning, in: Proceedings of the European Conference on Computer Vision (ECCV), 2025, pp. 311--328.

\bibitem{CSSL_challenges}
S.~Purushwalkam, P.~Morgado, A.~Gupta, The challenges of continuous self-supervised learning, in: Proceedings of the European Conference on Computer Vision (ECCV), 2022, pp. 702--721.

\bibitem{CSSL_cassle_Fini}
E.~Fini, V.~G.~T. da~Costa, X.~Alameda-Pineda, E.~Ricci, K.~Alahari, J.~Mairal, Self-supervised models are continual learners, in: Proceedings of the IEEE/CVF Conference on Computer Vision and Pattern Recognition (CVPR), 2022, pp. 9621--9630.

\bibitem{CSSL_C2ASR}
H.~Cheng, H.~Wen, X.~Zhang, H.~Qiu, L.~Wang, H.~Li, Contrastive continuity on augmentation stability rehearsal for continual self-supervised learning, in: Proceedings of the IEEE/CVF International Conference on Computer Vision (ICCV), 2023, pp. 5684--5694.

\bibitem{medical_CL_survey_1}
W.~Xinyao, X.~Zhe, T.~Raymond, Kai-yu, Continual learning in medical image analysis: {A} survey, Computers in Biology and Medicine 182 (2024) 109206.

\bibitem{medcoss}
Y.~Ye, Y.~Xie, J.~Zhang, Z.~Chen, Q.~Wu, Y.~Xia, Continual self-supervised learning: {T}owards universal multi-modal medical data representation learning, in: Proceedings of the IEEE/CVF Conference on Computer Vision and Pattern Recognition (CVPR), 2024, pp. 11114--11124.

\bibitem{tasai_icassp_2025}
R.~Tasai, G.~Li, R.~Togo, M.~Tang, T.~Yoshimura, H.~Sugimori, K.~Hirata, T.~Ogawa, K.~Kudo, M.~Haseyama, Continual self-supervised learning considering medical domain knowledge in chest {CT} images, in: Proceedings of the IEEE International Conference on Acoustics, Speech and Signal Processing (ICASSP), 2025, pp. 1--5.

\bibitem{CSSL_lump}
D.~Madaan, J.~Yoon, Y.~Li, Y.~Liu, S.~J. Hwang, Representational continuity for unsupervised continual learning, in: Proceedings of the International Conference on Learning Representations (ICLR), 2022, pp. 1--18.

\bibitem{CSSL_streaming_data_Hu}
D.~Hu, S.~Yan, Q.~Lu, L.~Hong, H.~Hu, Y.~Zhang, Z.~Li, X.~Wang, J.~Feng, How well does self-supervised pre-training perform with streaming data?, in: Proceedings of the International Conference on Learning Representations (ICLR), 2022, pp. 1--23.

\bibitem{catastrophic_forgetting_1993}
R.~M. French, Catastrophic interference in connectionist networks: {C}an it be predicted, can it be prevented?, in: Proceedings of the Advances in Neural Information Processing Systems (NeurIPS), Vol.~6, 1993, pp. 1176--1177.

\bibitem{catastrophic_forgetting_2017}
J.~Kirkpatrick, R.~Pascanu, N.~Rabinowitz, J.~Veness, G.~Desjardins, A.~A. Rusu, K.~Milan, J.~Quan, T.~Ramalho, A.~Grabska-Barwinska, et~al., Overcoming catastrophic forgetting in neural networks, Proceedings of the National Academy of Sciences 114~(13) (2017) 3521--3526.

\bibitem{experience_replay_1}
D.~Rolnick, A.~Ahuja, J.~Schwarz, T.~Lillicrap, G.~Wayne, Experience replay for continual learning, in: Proceedings of the Advances in Neural Information Processing Systems (NeurIPS), Vol.~32, 2019, pp. 350--360.

\bibitem{experience_replay_2}
P.~Buzzega, M.~Boschini, A.~Porrello, S.~Calderara, Rethinking experience replay: {A} bag of tricks for continual learning, in: Proceedings of the International Conference on Pattern Recognition (ICPR), 2021, pp. 2180--2187.

\bibitem{data_privacy_1}
A.~Ziller, D.~Usynin, R.~Braren, M.~Makowski, D.~Rueckert, G.~Kaissis, Medical imaging deep learning with differential privacy, Scientific Reports 11 (2021) 13524.

\bibitem{data_privacy_2}
B.~Sahiner, W.~Chen, R.~K. Samala, N.~Petrick, Data drift in medical machine learning: {I}mplications and potential remedies, The British Journal of Radiology 96~(1150) (2023) 20220878.

\bibitem{latent_replay_natural_1}
T.~L. Hayes, K.~Kafle, R.~Shrestha, M.~Acharya, C.~Kanan, Remind your neural network to prevent catastrophic forgetting, in: Proceedings of the European Conference on Computer Vision (ECCV), 2020, pp. 466--483.

\bibitem{SSL_CT_multi-domain_wolf2023sel}
D.~Wolf, T.~Payer, C.~S. Lisson, C.~G. Lisson, M.~Beer, M.~G{\"o}tz, T.~Ropinski, Self-supervised pre-training with contrastive and masked autoencoder methods for dealing with small datasets in deep learning for medical imaging, Scientific Reports 13 (2023) 20260.

\bibitem{SSL_CT_multi-domain_jiang2025self}
J.~Jiang, A.~Rangnekar, H.~Veeraraghavan, Self-supervised learning improves robustness of deep learning lung tumor segmentation models to {CT} imaging differences, Medical Physics 52~(3) (2025) 1573--1588.

\bibitem{tasai_gcce_2024}
R.~Tasai, G.~Li, R.~Togo, M.~Tang, T.~Yoshimura, H.~Sugimori, K.~Hirata, T.~Ogawa, K.~Kudo, M.~Haseyama, {Lung cancer classification using masked autoencoder pretrained on J-MID database}, in: Proceedings of the IEEE Global Conference on Consumer Electronics (GCCE), 2024, pp. 456--457.

\bibitem{SSL_MRI_multi-domain_chang2022self}
X.~Chang, X.~Cai, Y.~Dan, Y.~Song, Q.~Lu, G.~Yang, S.~Nie, Self-supervised learning for multi-center magnetic resonance imaging harmonization without traveling phantoms, Physics in Medicine \& Biology 67~(14) (2022) 145004.

\bibitem{SSL_MRI_multi-domain_fiorentino2024intensity}
M.~C. Fiorentino, F.~P. Villani, R.~Benito~Herce, M.~A. Gonz{\'a}lez~Ballester, A.~Mancini, K.~L{\'o}pez-Linares~Rom{\'a}n, An intensity-based self-supervised domain adaptation method for intervertebral disc segmentation in magnetic resonance imaging, International Journal of Computer Assisted Radiology and Surgery 19~(9) (2024) 1753--1761.

\bibitem{SSL_fundus_multi-domain_Gu2023102904}
R.~Gu, G.~Wang, J.~Lu, J.~Zhang, W.~Lei, Y.~Chen, W.~Liao, S.~Zhang, K.~Li, D.~N. Metaxas, S.~Zhang, {CDDSA}: {C}ontrastive domain disentanglement and style augmentation for generalizable medical image segmentation, Medical Image Analysis 89 (2023) 102904.

\bibitem{SSL_fundus_multi-domain_mojab2020real}
N.~Mojab, V.~Noroozi, D.~Yi, M.~P. Nallabothula, A.~Aleem, P.~S. Yu, J.~A. Hallak, Real-world multi-domain data applications for generalizations to clinical settings, in: Proceedings of the IEEE International Conference on Machine Learning and Applications (ICMLA), 2020, pp. 677--684.

\bibitem{DomainNet}
X.~Peng, Q.~Bai, X.~Xia, Z.~Huang, K.~Saenko, B.~Wang, Moment matching for multi-source domain adaptation, in: Proceedings of the IEEE International Conference on Computer Vision (ICCV), 2019, pp. 1406--1415.

\bibitem{parameter_regularization}
R.~Aljundi, F.~Babiloni, M.~Elhoseiny, M.~Rohrbach, T.~Tuytelaars, Memory aware synapses: {L}earning what (not) to forget, in: Proceedings of the European Conference on Computer Vision (ECCV), 2018, pp. 144--161.

\bibitem{retcop}
Y.~Yao, R.~Wu, Y.~Zhou, T.~Zhou, Continual retinal vision-language pre-training upon incremental imaging modalities, in: Proceedings of the International Conference on Medical Image Computing and Computer Assisted Intervention (MICCAI), 2025, pp. 111--121.

\bibitem{knowledge_distillation_survey_2025}
A.~M. Mansourian, R.~Ahmadi, M.~Ghafouri, A.~M. Babaei, E.~B. Golezani, Z.~Y. Ghamchi, V.~Ramezanian, A.~Taherian, K.~Dinashi, A.~Miri, S.~Kasaei, A comprehensive survey on knowledge distillation, Transactions on Machine Learning Research (2025).

\bibitem{knowledge_distillation_survey_2021}
J.~Gou, B.~Yu, S.~J. Maybank, D.~Tao, Knowledge distillation: {A} survey, International Journal of Computer Vision 129~(6) (2021) 1789--1819.

\bibitem{MAE}
K.~He, X.~Chen, S.~Xie, Y.~Li, P.~Doll{\'a}r, R.~Girshick, Masked autoencoders are scalable vision learners, in: Proceedings of the IEEE/CVF Conference on Computer Vision and Pattern Recognition (CVPR), 2022, pp. 16000--16009.

\bibitem{WKD_NeurIPS2024}
J.~Lv, H.~Yang, P.~Li, Wasserstein distance rivals {K}ullback-{L}eibler divergence for knowledge distillation, in: Proceedings of the Advances in Neural Information Processing Systems (NeurIPS), Vol.~37, 2024, pp. 65445--65475.

\bibitem{wkd_eq}
G.~Peyr'e, M.~Cuturi, Computational optimal transport, Foundations and Trends in Machine Learning 11~(5-6) (2019) 355--607.

\bibitem{wkd_diag}
E.~Yang, A.~C. Lozano, P.~Ravikumar, Elementary estimators for sparse covariance matrices and other structured moments, in: Proceedings of the International Conference on Machine Learning (ICML), 2014, pp. 397--405.

\bibitem{cov-2_dataset}
E.~Soares, P.~Angelov, S.~Biaso, M.~H. Froes, D.~K. Abe, {SARS-CoV-2 CT-scan dataset: A large dataset of real patients CT scans for SARS-CoV-2 identification}, MedRxiv (2020).

\bibitem{ricord_dataset}
E.~B. Tsai, S.~Simpson, M.~P. Lungren, M.~Hershman, L.~Roshkovan, E.~Colak, B.~J. Erickson, G.~Shih, A.~Stein, J.~Kalpathy-Cramer, et~al., {The RSNA international COVID-19 open radiology database (RICORD)}, Radiology 299~(1) (2021) E204--E213.

\bibitem{vit}
A.~Dosovitskiy, L.~Beyer, A.~Kolesnikov, D.~Weissenborn, X.~Zhai, T.~Unterthiner, M.~Dehghani, M.~Minderer, G.~Heigold, S.~Gelly, J.~Uszkoreit, N.~Houlsby, An image is worth 16x16 words: {T}ransformers for image recognition at scale, in: Proceedings of the International Conference on Learning Representations (ICLR), 2021, pp. 1--21.

\bibitem{bke}
Y.~Ge, C.~L. Choi, X.~Zhang, P.~Zhao, F.~Zhu, R.~Zhao, H.~Li, Self-distillation with batch knowledge ensembling improves {I}mage{N}et classification, arXiv preprint arXiv:2104.13298 (2021).

\bibitem{SSL_xray_li2023boosting}
G.~Li, R.~Togo, T.~Ogawa, M.~Haseyama, Boosting automatic {COVID-19} detection performance with self-supervised learning and batch knowledge ensembling, Computers in Biology and Medicine 158 (2023) 106877.

\bibitem{li2022self}
G.~Li, R.~Togo, T.~Ogawa, M.~Haseyama, Self-knowledge distillation based self-supervised learning for {COVID-19} detection from chest {X}-ray images, in: Proceedings of the IEEE International Conference on Acoustics, Speech and Signal Processing (ICASSP), 2022, pp. 1371--1375.

\bibitem{adam}
I.~Loshchilov, F.~Hutter, Decoupled weight decay regularization, in: Proceedings of the International Conference on Learning Representations (ICLR), 2019, pp. 1--18.

\end{thebibliography}

\end{document}